\documentclass[runningheads]{llncs}
\usepackage{graphicx}

\begin{document}
\title{Skill Check: Some Considerations on the Evaluation of Gamemastering Models for Role-playing Games}
\titlerunning{Some Considerations on the Evaluation of GM Models for RPGs}
\author{
Santiago Góngora\inst{1}\orcidID{0000-0002-6256-7577} \and
Luis Chiruzzo\inst{1}\orcidID{0000-0002-1697-4614} \and
Gonzalo Méndez\inst{2}\orcidID{0000-0001-7659-1482} \and
Pablo Gervás\inst{2}\orcidID{0000-0003-4906-9837}
}
\authorrunning{S. Góngora et al.}
\institute{Universidad de la República, Uruguay
\email{\{sgongora,luischir\}@fing.edu.uy}
\and
Universidad Complutense de Madrid, Spain
\email{\{gmendez,pgervas\}@ucm.es}
}% 
\maketitle              % 
\begin{abstract}
In role-playing games a Game Master (GM) is the player in charge of the game, who must design the challenges the players face and narrate the outcomes of their actions. 
In this work we discuss some challenges to model GMs from an Interactive Storytelling and Natural Language Processing perspective. 
Following those challenges we propose three test categories to evaluate such dialogue systems, and we use them to test ChatGPT, Bard and OpenAssistant as out-of-the-box GMs.

\end{abstract}
\section{Introduction}
Probably no one wants to hear somebody say ``\textit{Watch out! Behind that door there's a giant monster!}"; except if they are playing a role-playing game (RPG), using their imagination to visit endless worlds and having lots of fun.

Tabletop role-playing games (TTRPGs) consist of two or more players that collaborate in order to create a story, while acting as characters.
One of these players is the Game Master (GM), who is the one in charge of creating the world where the narrated events take place, describing the non-playable characters the human players meet and the situations they face. 
Having a player acting as the GM is one of the characteristics that most TTRPGs share \cite{UniversalGameEngine}.

Capturing the essence of RPGs has long been one of the goals of Interactive Storytelling (IS) research
\cite{PeinadoGervasGM,Riedl_Bulitko_2012}.
However, through the years only limited solutions have been found, typically by having a lot of premade scenes that can be mixed to generate other narrative structures\footnote{For example ``Call of Cthulhu: The Official Video Game'', an adaptation of the RPG.}, but pushing the player's freedom aside\footnote{An interesting example of this is ``The Stanley Parable'', a novel videogame that makes the players think about free will and the impact of their actions.} \cite{Riedl_Bulitko_2012}.
To automate a GM is a big challenge for Natural Language Processing (NLP) and Artificial Intelligence, due to its complexity on dialogue and creativity \cite{callison-burch-etal-2022-dungeons}.

Our long-term goal is to model 
the diverse set of 
skills that a GM needs to play RPGs.
This long path must lead to an explainable, grounded and controllable model, so human-in-the-loop features should be taken into consideration to meet the needs reported by \cite{RPGassistant} and \cite{zhu2023calypso}.

In this paper we will take a first step 
by proposing, inspired in core aspects of RPGs, a set of unit test categories  to evaluate such GM models. 
We also use these brand new tests to evaluate ChatGPT\footnote{https://openai.com/blog/chatgpt}, Bard\footnote{https://bard.google.com} and OpenAssistant\footnote{https://huggingface.co/chat/} \cite{kopf2023openassistant} as out-of-the-box automated GMs, both for Spanish and English.

\section{Previous work} \label{sec:previousWork}

The study of the role of GMs in narrative is nothing new and also some efforts have been made to explicitly model their capabilities. 
\cite{Tychsen2008TheMF} reflects on the core concepts of RPGs and gamemastering.
\cite{PeinadoGervasGM,formalization_GM_UML} consider GMs and RPGs as a framework to build IS systems, in order to tackle the \textit{Interactive Dilemma}, the conflict and balance between the player's will and the designer's choices.
Closely related to this concept is the GM's skill to improvise some aspects of a scene due to unexpected players' actions, 
and \cite{Martin2016,Martin2018DungeonsAD} discuss concepts, approaches and architectures taking that into account.
As they present insightful discussions about modelling narrative improvisation, they are some of the most clarifying works for us to date.

Most of the latest works pursue the modelling of GMs for \textit{Dungeons \& Dragons} (D\&D), called \textit{Dungeon Masters} (DMs), since it is the most popular RPG and finding data is easier than for other games. 
For example \cite{callison-burch-etal-2022-dungeons} describes the complexity of modelling the (D\&D) game, performing experiments with neural models and using \textit{control features} to guide their outputs. They also describe a \textit{gameplay dataset} in English used for training.
\cite{Theory-of-Mind-Dnd} tries to create a DM model with the ability to predict player's actions, modelling the \textit{Dungeon Master-Player} interactions using Theory-of-Mind and Reinforcement Learning.
The recent published datasets are also centered on modelling the D\&D game.
\cite{CritRoleDataset} presents one of the most complete datasets to study D\&D interactions, 
consisting of transcriptions of the popular \textit{Critical Role} web show. 
\cite{DeepDnD,zhu-etal-2023-fireball} also present datasets of D\&D players' interactions from online text-based playing sessions.

It is important to note that all of these recent works are about D\&D, while our main objective is to work on the general aspects of a GM, regardless of the specific game or theme. 
Additionally, all of them rely on English resources.

\section{A list of gamemastering challenges} \label{sec:requirements}

Most of the works mentioned in the previous section discuss difficulties faced while modelling some aspects of RPGs. 
However, as a way of introducing some details that guide our long-term goal and justify the test categories we propose, we would like to convey our thoughts on some challenges that a GM must face while running an RPG session. 
This list is not exhaustive and there may be other challenges that are not described here. 

\textbf{I. World and story design}. 
As storytellers, GMs must create and manage a rich and coherent world, populated with diverse forms of life (e.g. plants and animals) and characters.
In this fictional world is where the players' characters will live and act.
They also need to create some interesting places (e.g. an old library) and challenges for the players, which can be logic puzzles, tactic battles, complex dialogues with characters, or other challenges (e.g. the library has hidden rooms). Usually these situations are intended to be solved by teaming up with other characters, collaborating and using the different skills that they may master.
It is useful if a GM can also measure how interesting these challenges are for the players, and how meaningful they are for the development of their characters or other characters that live in the fictional world. 
That is the reason why it is important that such a model can take \textit{creative responsibility} \cite{BotelhoNonInteractiveStoryGen,CCFinalFrontier} while being able to explain what the plan
and objectives of each utterance are. 

\textbf{II. Extract player’s actions from input}. Since TTRPGs are played through a discussion between the players, these games have an inherent \textit{conversational nature}. Therefore classic research problems related to dialogue systems \cite{dialogue_survey} are fundamental to model GMs. 
More specifically, in order to \textit{understand} (i.e. semantically represent) the actions taken by the players, decide if they are possible in the fictional world and then determine the outcomes, the GM model should have the ability to semantically analyze their inputs.

\textbf{III. Commonsense reasoning}. Commonsense reasoning is an important research area within NLP \cite{sap-etal-2020-commonsense}, and despite the great advances made in the area it remains as one of the hardest tasks \cite{davis2015commonsense},
even for the recent Large Language Models (LLMs) like ChatGPT \cite{ChatGPTFeb2023}.
The relation between this classic task and the challenges for a GM is direct: since commonsense is an inherent part of our human identity, it naturally arises when playing RPGs. 
It is important to note that this challenge is related but different to the previous 
challenge: a model can semantically represent what a human is saying, but maybe the action does not make sense in some context. For instance, sometimes players may want to do actions that are possible in the real world but not in the fictional world (e.g. a character wants to play basketball but there is no gravity in her world).

\textbf{IV. Track the game state}. One of the core aspects of RPGs is to let the players act as they wish, what in IS is usually called \textit{user agency} \cite{Riedl_Bulitko_2012}. 
Making the players feel this way while thoroughly tracking the state of items (i.e. objects) and characters is one of the greatest problems for IS \cite{phdBenotti,Martin2016}. 
To track some component of the game is to know where it is, how hurt (in case of a living being) or damaged it is (in case of an object), and other properties that it may have (e.g. intensity of the magic property of a sacred object). This game state must be constantly updated as the world changes and the story moves forward.

Finally, we would like to mention other relevant aspects for this long path. 
In first place, we think it is crucial that the narrative structure and the game state may be represented using a human-readable format.
Since RPG games are used in educational \cite{gatsakou2021role} and therapy environments \cite{arenas2022therapeutic}, such GM models could be used to create serious games
with a wide range of objectives. 
Having the possibility of visualizing and customizing the boundaries of an RPG session is extremely crucial for that kind of applications. 

In second place, it is fundamental that these models generate respectful and \textit{ethical} outputs,
to make the players feel safe and included.
In modern RPGs like ``Alice is Missing''\footnote{https://www.huntersentertainment.com/alice-is-missing}
there are mechanics to silently communicate the rest of players that something recently said was hurtful or uncomfortable.
This is crucial when working with neural systems or LLMs, which are known to \textit{hallucinate} \cite{hallucination_survey} and generate offensive outputs \cite{StochasticParrots}.

Last but not least, we have to keep in mind that GMs are constantly adapting the game to fit the players' choices, so they have the additional requirement of facing every challenge described here on the fly.
That big challenge is related to what \cite{Martin2016} previously described as \textit{open-world improvisational storytelling}.

\section{How to evaluate such models?} \label{sec:experimentsdesign}

The procedure to evaluate creative systems (i.e. appropriate experiments and metrics) has long been a subject of debate, and remains one of the main problems of the field \cite{CCFinalFrontier,CC_Evaluation}. 
Since TTRPGs can be modelled as a series of utterances in a complex dialogue \cite{UniversalGameEngine}, we will assume that a GM model will always have a \textit{conversational nature},
as we mentioned in \textit{challenge II}. 
This gives us a general guideline: there is always a player who is asking or trying to do something, and another player answering or reacting to it.

The first idea that comes to mind could be to ask humans to play and evaluate the models based on their reaction. 
Although we consider important to measure how fun it is to play with the models, the humans' judgments can be very subjective, not very specific, and also biased by the fluency of the generated text \cite{behnke-etal-2022-bias}.
This bias can be stronger when working with LLMs, since they are trained to sound very natural to the human reader (exploiting the patterns behind the form of massive amounts of texts \cite{climbingNLU}), what can lead to distract the evaluators from their goal of judging specific characteristics of the models' output.

Hence, we would like to take an approach on evaluating basic, almost essential, skills that a GM should master.
We propose three different test categories related to the previously described challenges: \textit{commonsense reasoning}, the capacity to track \textit{items} in the world and the ability to coherently design \textit{maps}.
These categories were designed reflecting on core characteristics of RPGs, so we think they can be used to evaluate any system trying to model a GM, independently of the theme and features of the modelled game, and the technology used to play it.
We also hope these categories work as a guide for human evaluators, helping them to judge models
while reducing the subjectivity, the mentioned biases and the evaluation noise as possible. We will describe each of them next.

\subsection{GM-P-GM pattern}

In \textit{challenges II} and \textit{III} we discussed the importance of pragmatics and commonsense reasoning for a GM model. In order to evaluate the performance on this challenges we propose the GM-P-GM pattern, a formalization of the most elemental interaction between a GM and a player \cite{UniversalGameEngine}.
Specifically, we propose to evaluate the model's ability to judge the feasibility of a player's action:

\begin{itemize}
    \item $GM_1$: Narrates a \textbf{situation} to solve in some \textbf{context}.
    \item \textit{Player}: Describes the \textbf{actions} to overcome that \textbf{situation}.
    \item $GM_2$: Validates if those \textbf{actions} are feasible for that \textbf{context}, and next narrates the \textbf{outcomes}.
\end{itemize}

To run this test we give the model the $GM_1$ and \textit{Player} contradictory utterances and ask it to generate the $GM_2$ utterance. If the GM model prevents the action and explains why it is an inconsistency, the test is passed. A failure case is shown in table \ref{table_example_gm_p_gm}.

\begin{table}[h!]
\tiny
\centering
\begin{tabular}{|p{0.1\textwidth }|p{0.9\textwidth}|}
\hline
\textbf{GM} & You enter the building to look for the lost suitcase. However, the door is covered in flames. To your right you see a fire extinguisher, which seems to be ready to use. You can't use anything from your backpack because you can hardly move your arms. What will you do? \\ \hline
\textbf{Player} & I take a bucket of water from my backpack and put out the fire \\ \hline
\textbf{GM} & As the player grabs the bucket of water from their backpack, they realize it's not nearly enough to put out the flames completely. However, they quickly pour the water onto the fire, creating a small opening in the flames (...) \\ \hline
\textbf{Result} & Failed \\ \hline
\end{tabular}
  \caption{\label{table_example_gm_p_gm} An example of a \textit{GM-P-GM} unit test.}
\end{table}

\subsection{Item tracking}

As we described in \textit{challenge IV}, item tracking is one of the fundamental problems for gamemastering.
It is important both to track items that are located in specific places (e.g. on the ground, inside a safe) and to track items that belong to the characters.
The latter is usually referred as the \textit{inventory} of each character but, in order
to avoid the use of specific RPG words, in this test we will use \textit{bag} or \textit{backpack} as more neutral words for a similar concept.
To test if the model can track objects inside or outside the \textit{backpack}, we follow this scheme:

\begin{enumerate}
    \item \textit{Player}: ``What is in my backpack?''
    \item \textit{GM}: $Object_1, Object_2, Object_3, Object_4,...$
    \item \textit{Player}:  ``I take $Object_K$ and leave it on *another location*''
    \item \textit{GM}: *Narrates outcome*
    \item \textit{Player}:   ``What is in my backpack?''
    \item \textit{GM}: $Object_1, Object_2, Object_3,...$
    \item \textit{Player}:  ``What is on *another location* ?''
    \item \textit{GM}: $Object_A, Object_B,...$
\end{enumerate}

If the lists given in steps $\#2$ and $\#6$ are the same or contain other unrelated objects, then the test is failed. 
In case the object is not listed in step $\#6$, the step $\#8$ tries to check if the model updated its new location. A failure case is shown in table \ref{table_example_items}.

\begin{table}[h!]
\tiny
\centering
\begin{tabular}{|p{0.1\textwidth }|p{0.9\textwidth}|}
\hline
\textbf{GM}     & (...) Aside from the mysterious key, your backpack also contains the usual adventurer's equipment: a sturdy rope, a grappling hook, a water flask, some dried rations, a small first-aid kit, a compass, a map of the surrounding area, and a trusty dagger at your side.  \\ \hline
\textbf{Player} & I take the rope, the map and the key and I leave them on the floor \\ \hline
\textbf{GM}     & You carefully set the rope, map, and key down on the floor, making sure they are within reach in case you need them later. (...) \\ \hline
\textbf{Player} & What's in my backpack? \\ \hline
\textbf{GM}     & As you take a closer look at your backpack, you notice that it contains a few essential items that any adventurer would find useful:\\ 
& A sturdy rope: (...) this rope is perfect for climbing or rappelling down steep surfaces. (...) \\ 
& A map of the surrounding area: this map shows the nearby landmarks, towns, and topography of the region, allowing you to plan your route. (...) \\ \hline
\textbf{Result} & Failed  \\ \hline
\end{tabular}
\caption{\label{table_example_items} An example of an \textit{Item tracking} unit test.}
\end{table}

\subsection{Map design}

When GMs create a fictional world they also have the challenge to design an interesting and coherent concrete map for the players to explore, as we described in \textit{challenge I}. Although we also justified the importance of having interesting situations for the players, for now we do not focus on how attractive, large or complex the maps generated are. 

Inspired in the scheme of the \textit{Item tracking} category, we propose to ask the model to describe the accessible locations from the current location. Then the player must choose to move to one of those locations and go back again to the original location, in order to test the map's coherence. If during this process the available locations are coherent, the test is passed.  A successful case is shown in table \ref{table_example_map}.

\begin{table}[h!]
\tiny
\centering
\begin{tabular}{|p{0.1\textwidth }|p{0.9\textwidth}|}
\hline
\textbf{GM}     & (...) Current place: You find yourself standing at the entrance of a dark and damp cave. The entrance is surrounded by tall trees, and the sound of running water can be heard in the distance.\\ 
& Accessible places: Inside the cave, Back to the forest, Along the river \\ \hline
\textbf{Player} & I want to explore the cave   \\ \hline
\textbf{GM}     & (...) Current place: Inside the cave\\
& Accessible places: The cave entrance, Deeper into the cave \\ \hline
\textbf{Player} & I want to go outside    \\ \hline
\textbf{GM}     & (...) Current place: Outside the cave\\ 
& Accessible places: Inside the cave, Along the river, Deeper into the forest\\ \hline
\textbf{Result} & Passed \\ \hline
\end{tabular}
\caption{\label{table_example_map} An example of a \textit{Map design} unit test.}
\end{table}

\section{Large Language Models as out-of-the-box GMs} \label{sec:results}

In order to perform some experiments with our three proposed categories, we wrote five unit tests in Spanish and English for each of them.
We then used those unit tests to evaluate the out-of-the-box GM skills of three LLMs: ChatGPT\footnote{``ChatGPT Mar 14 Version. Free Research Preview''},
Bard\footnote{Bard Experimental. Accessed on the 17th of July, 2023.}
and the LLaMA-based OpenAssistant chatbot\footnote{``Model: OpenAssistant/oasst-sft-6-llama-30b''. Accessed on the 14th of July, 2023.}.
We consider these three LLMs appropriate 
since they are dialogue systems (i.e. they have a \textit{conversational nature}), a fundamental requirement for a GM model.
Since these are only preliminary experiments we consider really important to make the experimental logs open, because it can help the reader to critically examine the results reported here and reflect on the real flaws and strengths of both our proposed test categories and the evaluated models.
Therefore, the detailed logs of the experiments and their results
are accessible on GitHub\footnote{https://github.com/sgongora27/skill-check-GM-tests}.

We detail the analysis in two subsections, the quantitative results and the qualitative observations. 

\subsection{Quantitative results}

After we ran the tests using the aforementioned models, we carefully examined the outputs and determined the results for each test,
shown in table \ref{table_results}.
As can be seen, the performance on the \textit{GM-P-GM} category is really low, regardless of the language or model. 
This result is aligned with those in \cite{ChatGPTFeb2023}, where commonsense reasoning was one of the remarkable flaws of ChatGPT. However, \textit{Item tracking} and \textit{Map design} tests were quite good both for ChatGPT and Bard.
Although these preliminary experiments do not report a big gap in the results for ChatGPT or Bard between languages, they do unveil their strength over OpenAssistant. 
In most cases OpenAssistant just could not finish the test, due to generating nonsensical outputs that had nothing to do with the narrated events. 
That problem was even deeper for the tests in Spanish.

\begin{table*}[t!]
\centering
\footnotesize
\begin{tabular}{|l|l|l|l||l|l|l|}
\hline
\textbf{Category} & \textbf{OA} {[}ES{]} & \textbf{BARD} {[}ES{]} & \textbf{CGPT} {[}ES{]} & \textbf{OA} {[}EN{]} & \textbf{BARD} {[}EN{]} & \textbf{CGPT} {[}EN{]} \\ \hline
GM-P-GM                                   & 0 out of 5  & 1 out of 5    & 1 out of 5    & 1 out of 5  & 1 out of 5    & 0 out of 5    \\ \hline
Item                                     & 0 out of 5  & 0 out of 5    & 2 out of 5    & 0 out of 5  & 3 out of 5    & 1 out of 5    \\ \hline
Map                                       & 0 out of 5  & 3 out of 5    & 3 out of 5    & 0 out of 5  &  2 out of 5   & 3 out of 5    \\ \hline\hline
\textbf{Total} & 0 out of 15 & 4 out of 15 & 6 out of 15 & 1 out of 15 & 6 out of 15 & 4 out of 15 \\ \hline
\end{tabular}
  \caption{\label{table_results} Number of passed tests for each of the categories described in section \ref{sec:experimentsdesign}, testing OpenAssistant (OA), Google's Bard and ChatGPT (CGPT), both for English and Spanish. The last row shows the sum of the passed tests for each model-language pair.}
\end{table*}

\subsection{Qualitative observations}

The first and most important observation is that
ChatGPT and Bard are really good at making the user feel that is playing with a real GM. 
There is a world to interact with, characters to meet and items to use. 
Everything seems perfect if the player chooses an action from those suggested by the model, although it is far from perfect when having to improvise new scenes and keep it coherent.
OpenAssistant, however, struggles to deliver a minimum interactive experience and the tests had to be repeated several times to obtain a reasonable output. 
Our evaluation methodology does not distinguish that kind of errors, hence this aspect cannot be inferred by just comparing the quantitative results for each category between models (e.g. Bard failed the Spanish \textit{Item tracking} tests due to wrongly list the available items, while OpenAssistant failed them because could not even give a proper output).
However, we think that the quantitative results do represent the strengths and weaknesses of each model (e.g. ChatGPT is better at world coherence than commonsense reasoning) but also the ``Total'' scores provide an accurate comparison of the experience provided by the different models.

The second observation is about the contents generated by the models when taking the \textit{creative responsibility}. 
Almost every scene \textit{generated} by the models
took place in a medieval-fantasy setting.
This relation between RPGs and a medieval setting is aligned with the previous comments in section \ref{sec:previousWork}: most of the available data about RPGs is in fact about D\&D.
As LLMs reproduce the biases in their training data \cite{StochasticParrots},
this shows that more work on other RPGs with different themes is needed.
There is also a notorious absence of diversity of plots; after playing a few hours the narrated events and the available places start to repeat.
Although this is related to the previous comment about the medieval settings and the biased data, it is important to have in mind that a great diversity of plots can be created using a medieval-fantasy setting\footnote{This is evidenced by the massive amount of adventures published for RPGs with this theme, such as \textit{Dungeons \& Dragons} or \textit{Pathfinder}.}, so they are independent flaws and might be studied separately.

Our third observation is about these models' tendency to constantly adjust the output to the prompt. 
If the player says or tries to do something
then the output will try to adjust the narrative to it, without letting the player to feel any mystery about the plot. 
This is not a good sign for the
skills we described in \textit{challenge I}.

\subsection{Limitations}

Although we propose the test categories to assist the evaluation of GM models, 
the human subjectivity is still there.
In addition to the difficulties faced to decide whether or not the test was passed, this subjectivity can also be present in the prompts design as well, as in the case of the \textit{GM-P-GM} tests which need a specific human-designed case in order to run (i.e. a situation to solve and a player's solution to it).  
To perform a deeper evaluation and extract stronger conclusions we would need a diverse team of human evaluators and a bigger number of tests.

It is important to highlight that the difficulties faced when evaluating a creative system, added to the nearly-infinite input space that RPGs offer, make the evaluation even harder. 
Furthermore, the LLMs show a tendency to irregularly move the story forward: sometimes the model's output narrates a single event happening immediately and sometimes narrates long scenes.
Not having a reliable mechanism (e.g. a symbolic representation) to restrict the model makes the execution of these tests more unpredictable, forcing the human evaluator to take unexpected decisions on the fly.
For example, it would be positive for the \textit{Map design} tests to have some kind of constraints and visualization components, to perform an in-depth analysis of the different reachable places in a given scene but without moving the story forward.
Also, these dialogue models compute the utterances each time a new input is sent, what makes the replication experiments harder. 
Additionally we share the same limitations found by \cite{ChatGPTFeb2023} regarding the needed time to run a small set of tests.

\section{Conclusions and future work} \label{sec:conclusions}

In this paper we discussed some challenges to face in order to model the skills that GMs need to play RPGs, like creating and managing a fictional world, tracking the game state and understanding the players' actions.

Following those challenges we proposed three test categories to evaluate any kind of GM model. Although these tests are domain specific, we think they can inspire other evaluation methodologies for dialogue systems.

We also used those test categories to perform preliminary experiments with ChatGPT, Bard and OpenAssistant.
We found that ChatGPT and Bard can provide a satisfying gaming experience, but also they struggle when dealing with commonsense reasoning.
OpenAssistant was unable to maintain the GM role during most of the tests. 
All 90 unit tests are available on GitHub.

The difficulties faced to control the models' outputs while running the tests make us think that in the future more \textit{neuro-symbolic} approaches should be explored.
We think that would help to keep the test phase more controllable, and also allow the players to examine the narrative details, avoid some scenes that they do not want to play and add another elements that they do.

In the future we would like to improve these test categories and design more to test other gamemastering skills (e.g. model the emotional variation of a character during an interaction with other character \cite{emolift}).

\section{Acknowledgements}
This paper has been partially funded by ANII (Uruguayan Innovation and Research National Agency), Grant No. $POS\_NAC\_2022\_1\_173659$ and by the project CANTOR: Automated Composition of Personal Narratives as an aid for Occupational Therapy based on Reminescence, Grant No. $PID2019-108927RB-I00$ (Spanish Ministry of Science and Innovation).

\bibliographystyle{splncs04}
\bibliography{bibliography}

\begin{thebibliography}{10}
\providecommand{\url}[1]{\texttt{#1}}
\providecommand{\urlprefix}{URL }
\providecommand{\doi}[1]{https://doi.org/#1}

\bibitem{RPGassistant}
Acharya, D., Mateas, M., Wardrip-Fruin, N.: Story improvisation in tabletop
  roleplaying games: Towards a computational assistant for game masters. In:
  2021 IEEE Conference on Games (CoG). pp. 01--08 (2021).
  \doi{10.1109/CoG52621.2021.9619006}

\bibitem{arenas2022therapeutic}
Arenas, D.L., Viduani, A., Araujo, R.B.: Therapeutic use of role-playing game
  (rpg) in mental health: A scoping review. Simulation \& Gaming
  \textbf{53}(3),  285--311 (2022)

\bibitem{behnke-etal-2022-bias}
Behnke, H., Fomicheva, M., Specia, L.: Bias mitigation in machine translation
  quality estimation. In: Proceedings of the 60th Annual Meeting of the
  Association for Computational Linguistics (Volume 1: Long Papers). pp.
  1475--1487. Association for Computational Linguistics, Dublin, Ireland (May
  2022). \doi{10.18653/v1/2022.acl-long.104}

\bibitem{StochasticParrots}
Bender, E.M., Gebru, T., McMillan-Major, A., Shmitchell, S.: On the dangers of
  stochastic parrots: Can language models be too big? In: Proceedings of the
  2021 ACM Conference on Fairness, Accountability, and Transparency. p.
  610–623. FAccT '21, Association for Computing Machinery, New York, NY, USA
  (2021). \doi{10.1145/3442188.3445922}

\bibitem{climbingNLU}
Bender, E.M., Koller, A.: Climbing towards {NLU}: {On} meaning, form, and
  understanding in the age of data. In: Proceedings of the 58th Annual Meeting
  of the Association for Computational Linguistics. pp. 5185--5198. Association
  for Computational Linguistics, Online (Jul 2020).
  \doi{10.18653/v1/2020.acl-main.463}

\bibitem{phdBenotti}
Benotti, L.: Implicature as an Interactive Process. (L'implicature comme un
  Processus Interactif). Ph.D. thesis, Henri Poincar{\'{e}} University, Nancy,
  France (2010)

\bibitem{BotelhoNonInteractiveStoryGen}
Botelho, L.M.: A guided journey through non-interactive automatic story
  generation (2021)

\bibitem{callison-burch-etal-2022-dungeons}
Callison-Burch, C., Tomar, G.S., Martin, L., Ippolito, D., Bailis, S., Reitter,
  D.: Dungeons and dragons as a dialog challenge for artificial intelligence.
  In: Proceedings of the 2022 Conference on Empirical Methods in Natural
  Language Processing. pp. 9379--9393. Association for Computational
  Linguistics, Abu Dhabi, United Arab Emirates (Dec 2022)

\bibitem{dialogue_survey}
Chen, H., Liu, X., Yin, D., Tang, J.: A survey on dialogue systems: Recent
  advances and new frontiers. SIGKDD Explor. Newsl.  \textbf{19}(2),  25–35
  (nov 2017). \doi{10.1145/3166054.3166058}

\bibitem{CCFinalFrontier}
Colton, S., Wiggins, G.A.: Computational creativity: The final frontier? In:
  European Conference on Artificial Intelligence (2012)

\bibitem{davis2015commonsense}
Davis, E., Marcus, G.: Commonsense reasoning and commonsense knowledge in
  artificial intelligence. Communications of the ACM  \textbf{58}(9),  92--103
  (2015)

\bibitem{UniversalGameEngine}
Durall, J., Perrin, S.: Basic Roleplaying: Universal Game Engine. Chaosium Inc.
  (2023)

\bibitem{gatsakou2021role}
Gatsakou, C., Bardis, N., Drigas, A.: Role playing vs rpgs as teaching
  strategies in educational procedure. Technium Soc. Sci. J.  \textbf{26}, ~186
  (2021)

\bibitem{hallucination_survey}
Ji, Z., Lee, N., Frieske, R., Yu, T., Su, D., Xu, Y., Ishii, E., Bang, Y.J.,
  Madotto, A., Fung, P.: Survey of hallucination in natural language
  generation. ACM Comput. Surv.  \textbf{55}(12) (mar 2023).
  \doi{10.1145/3571730}

\bibitem{CC_Evaluation}
Jordanous, A.: A standardised procedure for evaluating creative systems:
  Computational creativity evaluation based on what it is to be creative. Cogn.
  Comput.  \textbf{4}(3),  246--279 (2012). \doi{10.1007/s12559-012-9156-1}

\bibitem{kopf2023openassistant}
Köpf, A., Kilcher, Y., von Rütte, D., Anagnostidis, S., Tam, Z.R., Stevens,
  K., Barhoum, A., Duc, N.M., Stanley, O., Nagyfi, R., ES, S., Suri, S.,
  Glushkov, D., Dantuluri, A., Maguire, A., Schuhmann, C., Nguyen, H., Mattick,
  A.: Openassistant conversations -- democratizing large language model
  alignment (2023)

\bibitem{DeepDnD}
Louis, A., Sutton, C.: Deep dungeons and dragons: Learning character-action
  interactions from role-playing game transcripts. In: Proceedings of the 2018
  Conference of the North {A}merican Chapter of the Association for
  Computational Linguistics: Human Language Technologies, Volume 2 (Short
  Papers). pp. 708--713. Association for Computational Linguistics, New
  Orleans, Louisiana (Jun 2018). \doi{10.18653/v1/N18-2111}

\bibitem{Martin2016}
Martin, L.J., Harrison, B., Riedl, M.O.: Improvisational computational
  storytelling in open worlds. In: Nack, F., Gordon, A.S. (eds.) Interactive
  Storytelling. pp. 73--84. Springer International Publishing, Cham (2016)

\bibitem{Martin2018DungeonsAD}
Martin, L.J., Sood, S., Riedl, M.O.: Dungeons and dqns: Toward reinforcement
  learning agents that play tabletop roleplaying games. In: INT/WICED@AIIDE
  (2018)

\bibitem{emolift}
O{\~n}ate, A., M{\'e}ndez, G., Gerv{\'a}s, P.: Emolift: Elevator conversations
  based on emotions. In: International Conference on Innovative Computing and
  Cloud Computing (2019)

\bibitem{PeinadoGervasGM}
Peinado, F., Gerv{\'a}s, P.: Transferring game mastering laws to interactive
  digital storytelling. In: G{\"o}bel, S., Spierling, U., Hoffmann, A., Iurgel,
  I., Schneider, O., Dechau, J., Feix, A. (eds.) Technologies for Interactive
  Digital Storytelling and Entertainment. pp. 48--54. Springer Berlin
  Heidelberg, Berlin, Heidelberg (2004)

\bibitem{formalization_GM_UML}
Peinado, F., Gerv{\'a}s, P.: Automatic direction of interactive storytelling:
  Formalizing the game master paradigm. In: Cavazza, M., Donikian, S. (eds.)
  Virtual Storytelling. Using Virtual Reality Technologies for Storytelling.
  pp. 196--201. Springer Berlin Heidelberg, Berlin, Heidelberg (2007)

\bibitem{ChatGPTFeb2023}
Qin, C., Zhang, A., Zhang, Z., Chen, J., Yasunaga, M., Yang, D.: Is chatgpt a
  general-purpose natural language processing task solver? (2023).
  \doi{10.48550/ARXIV.2302.06476}

\bibitem{CritRoleDataset}
Rameshkumar, R., Bailey, P.: Storytelling with dialogue: {A} {Critical Role
  Dungeons and Dragons Dataset}. In: Proceedings of the 58th Annual Meeting of
  the Association for Computational Linguistics. pp. 5121--5134. Association
  for Computational Linguistics, Online (Jul 2020).
  \doi{10.18653/v1/2020.acl-main.459}

\bibitem{Riedl_Bulitko_2012}
Riedl, M.O., Bulitko, V.: Interactive narrative: An intelligent systems
  approach. AI Magazine  \textbf{34}(1), ~67 (Dec 2012).
  \doi{10.1609/aimag.v34i1.2449}

\bibitem{sap-etal-2020-commonsense}
Sap, M., Shwartz, V., Bosselut, A., Choi, Y., Roth, D.: Commonsense reasoning
  for natural language processing. In: Proceedings of the 58th Annual Meeting
  of the Association for Computational Linguistics: Tutorial Abstracts. pp.
  27--33. Association for Computational Linguistics, Online (Jul 2020).
  \doi{10.18653/v1/2020.acl-tutorials.7}

\bibitem{Tychsen2008TheMF}
Tychsen, A., Hitchens, M.: The many faces of role-playing games. International
  journal of role-playing  \textbf{1}(1),  3--21 (2009)

\bibitem{Theory-of-Mind-Dnd}
Zhou, P., Zhu, A., Hu, J., Pujara, J., Ren, X., Callison-Burch, C., Choi, Y.,
  Ammanabrolu, P.: An ai dungeon master's guide: Learning to converse and guide
  with intents and theory-of-mind in dungeons and dragons (2022).
  \doi{10.48550/ARXIV.2212.10060}

\bibitem{zhu-etal-2023-fireball}
Zhu, A., Aggarwal, K., Feng, A., Martin, L., Callison-Burch, C.: {FIREBALL}: A
  dataset of dungeons and dragons actual-play with structured game state
  information. In: Proceedings of the 61st Annual Meeting of the Association
  for Computational Linguistics (Volume 1: Long Papers). pp. 4171--4193.
  Association for Computational Linguistics, Toronto, Canada (Jul 2023)

\bibitem{zhu2023calypso}
Zhu, A., Martin, L.J., Head, A., Callison-Burch, C.: Calypso: Llms as dungeon
  masters' assistants. In: AAAI Conference on Artificial Intelligence and
  Interactive Digital Entertainment (AIIDE). AAAI, Salt Lake City, UT (10 2023)

\end{thebibliography}

\end{document}